\def\year{2020}\relax
\documentclass[letterpaper]{article} 
\usepackage{aaai20}  
\usepackage{times}  
\usepackage{helvet} 
\usepackage{courier}  
\usepackage[hyphens]{url}  
\usepackage{graphicx} 
\usepackage{graphicx}  
\usepackage{times}
\usepackage{helvet}
\usepackage{courier}
\usepackage{color}
\usepackage{graphicx}
\usepackage{amsmath,amssymb,amsfonts}
\usepackage[ruled,vlined,linesnumbered,boxed,noend]{algorithm2e}
\usepackage{hyperref} 
\usepackage{cleveref}
\usepackage{balance}
\usepackage{booktabs}
\usepackage{soul}

\begin{document}
%
\setlength\titlebox{2.5in} 
\title{
A Dynamic Deep Neural Network For Multimodal Clinical Data Analysis\\
}
\author{Maria H\"ugle$^{1}$, Gabriel Kalweit$^{1}$, Thomas H\"ugle$^{2}$ and Joschka Boedecker$^{1,3}$\\
$^{1}$Neurorobotics Lab, Department of Computer Science, University of Freiburg, Germany.\\
$^{2}$Department of Rheumatology, University Hospital Lausanne, CHUV, Switzerland.\\
$^{3}$Cluster of Excellence BrainLinks-BrainTools, University of Freiburg, Germany.\\
\{hueglem, kalweitg, jboedeck\}@cs.uni-freiburg.de, Thomas.Hugle@chuv.ch
}

\maketitle
\begin{abstract}
\begin{quote}
Clinical data from electronic medical records, registries or trials provide a large source of information to apply machine learning methods in order to foster precision medicine, e.g. by finding new disease phenotypes or performing individual disease prediction. However, to take full advantage of deep learning methods on clinical data, architectures are necessary that 1) are robust with respect to missing and wrong values, and 2) can deal with highly variable-sized lists and long-term dependencies of individual diagnosis, procedures, measurements and medication prescriptions. 
In this work, we elaborate limitations of fully-connected neural networks and classical machine learning methods in this context and propose AdaptiveNet, a novel recurrent neural network architecture, which can deal with multiple lists of different events, alleviating the aforementioned limitations. We employ the architecture to the problem of disease progression prediction in rheumatoid arthritis using the Swiss Clinical Quality Management registry, which contains over 10.000 patients and more than 65.000 patient visits. Our proposed approach leads to more compact representations and outperforms the classical baselines.
\end{quote}
\end{abstract}

\section{Introduction}
Driven by increased computational power and larger datasets, deep learning (DL) techniques have successfully been applied to process and understand complex data~\cite{LeCunDL}. In recent years, the adoption of electronic medical records (EMRs), registries and trial datasets has heavily increased the amount of captured patient data from thousands up to millions of individuals patients. Those datasets can be used to predict individual disease progression and outcomes in medicine to assist patients and doctors. Especially deep learning methods are more and more applied to process clinical data \cite{Miotto2016DeepPA,DBLP:journals/corr/ShickelTBR17,Rajkomar,HOANG2019104788} and learn from former experiences on a large scale as a potential tool to guide treatment and surveillance \cite{Komorowski}.\\

To establish high-quality decision making systems for clinical datasets, machine learning (ML) methods have to be able to deal with the varying structure of the data, containing variable-sized lists of diagnosis, procedures, measurements and medication prescriptions. Capturing long-term dependencies and handling irregular event sequences with variable time spans in between is crucial to model complex disease progressions of patients in personalized medicine.\\

\begin{figure}[t]
    \centering
    \includegraphics[width=0.45\textwidth]{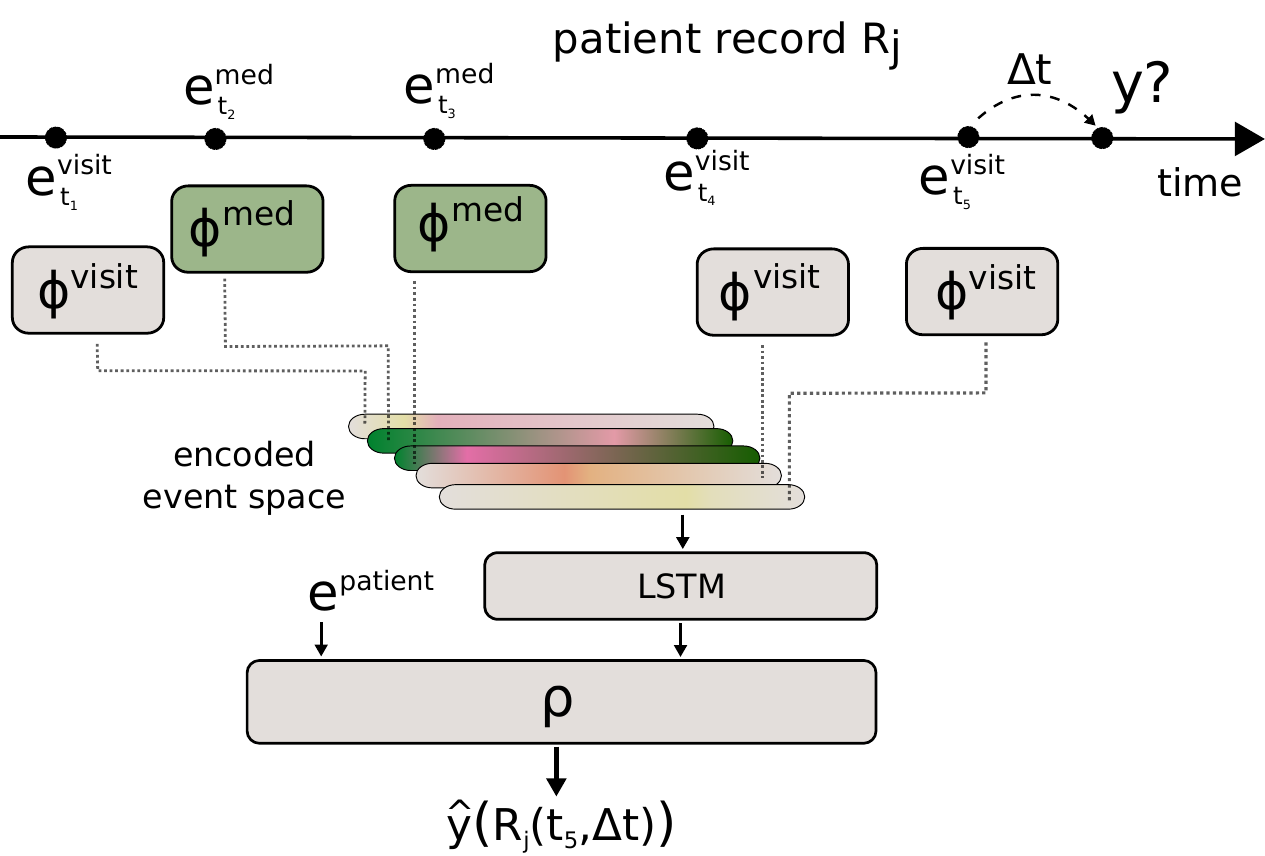}
    \caption{Scheme of AdaptiveNet, which projects visits and medication adjustments to the same latent space using encoder networks $\phi^{\text{visit}}$ and $\phi^{\text{med}}$, where the output vectors $\phi^{(\cdot)}{(\cdot)}$ have the same length. The sorted list of encoded events are pooled by an LSTM to compute a fixed-length encoded patient history. The final output $\hat y$ is computed by the network module  $\rho$.}
    \label{fig:adaptivenet}
\end{figure}

Classical ML models, such as Random Forests (RFs) \cite{breiman2001random}, and  fully-connected networks (FCNs) or recurrent neural networks (RNNs) were applied successfully to predict disease progression \cite{Vodencarevic} and disease outcomes \cite{KOUROU20158}. However, classical FCN architectures\footnote{As architecture we understand the full specification of interactions between different network modules, in- and outputs. In the following, we denote classical fully-connected neural networks without any extensions as FCN.} are limited to a fixed number of input features. As consequence, these models can only consider a fixed number of events, such as visits or medication prescriptions and adjustments. As a simple workaround, only the last $N$ visits and $M$ medication prescriptions can be considered and older entries ignored. In case patients have less than $N$ entries for visits or $M$ medication adjustments, dummy values have to be used to guarantee a fixed number of input features. However, depending on the choice of $N$ and $M$, the full patient history is not considered.\\

Classical recurrent neural networks, such as long short-term memories (LSTMs) \cite{Hochreiter:1997:LSM:1246443.1246450} can deal with variable-sized inputs and can handle irregularly timed events. In previous work, LSTMs were used for outcome prediction in intensive care units \cite{Lipton2015LearningTD}, heart failure prediction \cite{Maragatham:2019:LMP:3324573.3324793}, Alzheimer's disease progression prediction \cite{LeeAlzheimer} and other applications \cite{Baytas:2017:PSV:3097983.3097997,PHAM2017218}.  However, the proposed architectures can only deal with one variable-sized list of one feature representation, which is limiting when working with clinical data. In this work, we aim at exploiting and unifying all available patient data. The resulting neural network architecture has to be able to deal with multiple variable-sized lists of different events like visits, medication adjustments or imaging, which all have different feature representations. In \cite{DBLP:journals/corr/abs-1803-04837,DBLP:journals/corr/abs-1903-08652}, this problem was approached by using large one-hot encodings, which contain features for all different event types. However, this approach leads to a high amount of dummy values and has limited flexibility.\\

We propose AdaptiveNet, a recurrent neural network architecture that can be trained in an end-to-end fashion. The scheme of the architecture is shown in \Cref{fig:adaptivenet}. AdaptiveNet can deal with multiple variable-sized lists of different event types by using multiple fully-connected encoding network modules to project all event types to the same latent space. Then, the sorted list of latent event representations is fed to a recurrent unit in order to compute a latent representation describing the full patients history. We employ AdaptiveNet for the problem of disease progression prediction in rheumatoid arthritis (RA) and evaluate the performance on the Swiss Clinical Quality Management (SCQM) dataset.

\section{Related Work}

Data in health care and biology often contains a variety of missing and wrong values. Previous work already addressed missing values in time series \cite{CIS-195921}, but did not provide good performance when the missing rate was too high and inadequate samples were kept. While classical methods can omit missing data by performing analysis only on the observed data, for deep neural network architectures this is not straightforward. A possible solution is to fill in the missing values with substituted values, e.g. by smoothing or interpolation  \cite{Nancy:2017:IMV:3096755.3096890}. In \cite{InferenceMissingData}, it was shown  that missing values can provide useful information about target labels in supervised learning tasks. In \cite{YanLiu} this fact was exploited by using recurrent neural networks that are aware of missing values.\\

Other network architectures, such as convolutional neural networks (CNNs) \cite{nguyen2016deepr}, Transformer architectures \cite{DBLP:journals/corr/abs-1907-09538} or graph neural networks (GNNs) \cite{DBLP:journals/corr/abs-1812-08434} were also used to process clinical data \cite{DBLP:journals/corr/abs-1906-04716}. In contrast to CNNs, which are limited to their initial grid size, Transformers and GNNs are able to deal with variable-sized input lists. In general, these architectures could be extended to deal with multiple variable-sized lists in the same manner as described in this work for RNNs. However, since we want to cover temporal long-term dependencies in timeseries, we focus on RNNs.\\

Projecting objects to a latent space and pooling the latent vectors was already proposed in the Deep Set architecture \cite{NIPS2017_6931}. As pooling component, any permutation-invariant operator can be used, such as the \textit{sum}. This approach was extended in \cite{huegle2019dynamic} by projecting different object types to the same latent space using multiple encoders and pooling the objects by the \textit{sum} in the context of off-policy reinforcement learning. In this work, we use a recurrent unit as pooling function to cover temporal dependencies.\\

In the application of RA, there is only few prior work on disease progression prediction. In \cite{Vodencarevic}, rheumatic flares were predicted using Logistic Regression and Random Forests. Defining flares by DAS28-EST $\ge$ 3.2 and Swollen Joint Count 28 $\ge$ 2, they achieved an AUC of about 80\% in a small study with a group of 314 carefully selected patients from the Utrecht Patient Oriented Database. 
Disease detection, which is less sophisticated than disease progression prediction, was performed in~\cite{Shiezadeh}. Using an ensemble of decision trees, they achieved an accuracy of 85\% and a sensitivity and specificity of 44\% and 74\% on a dataset of around 2500 patients referred to a rheumatology clinic in Iran. \cite{Lin} performed disease detection based on classical ML methods trained on  $>2500$ clinical notes and lab values. With an SVM, they achieved the best performance with an AUC score of 0.831.

\section{Methods}
In this section, we describe the disease progression prediction problem and how input samples can be generated from clinical datasets. Finally, the architecture of AdaptiveNet is explained in detail.

\subsection{Disease Progression Prediction}

Disease progression prediction based on a clinical dataset can be modeled as time-series prediction, where for a patient at time point $t$ the future disease activity at time point $ t + \Delta t$ is predicted. The dataset consists of records for a set of patients $\mathcal{R} = \bigcup_j \mathcal{R}_j$. 
Records can contain general patient information (e.g. age, gender, antibody-status) and multiple list of events. The subset $R_j(t) \subseteq \mathcal{R}_j$, denotes all records of a patient  $j$  collected until time point $t$ with
$$R_j(t) = \{e^\text{patient}(t)\} \cup E^\text{visit}(t) \cup E^{\text{med}}(t),$$
where $e^\text{patient}(t)$ contains general patient information collected until time point $t$. $E^{k}(t)$ is a list of events collected until time point $t$, where
$$E^{k}(t) = \{e^{k}(t_e)\  | \ e^{k}(t_e) \in \mathcal{R}_j \text{ and } t_e \le t \} ,$$ for all event types $k \in \{\text{visit}, \text{med}\}$. Visit events contain information like joint swelling or patient reported outcomes (e.g. joint pain, morning stiffness and HAQ) and lab values (e.g. CRP, BSR). Medication events contain adjustment information, such as drug type and dose. In the same manner, further lists of other event types could be added, such as imaging data (e.g. MRI, radiograph). Considering records $R_j(t)$ as input, we aim to learn a function $ f: (R_j(t), \Delta t) \rightarrow \mathbb{R}$ that maps the records at time point $t$ to the expected change of the disease level $\text{score}_j$ until time $t + \Delta t$ with
$$f( R_j(t), \Delta t) =  \text{score}_j(t + \Delta t) - \text{score}_j(t).$$
To account for variable time spans, for all events, the time distance $\Delta t$ to the prediction time point is added as additional input feature. Records with included time feature are denoted as $R_j(t, \Delta t)$ and event lists with $E^{(\cdot)}(t, \Delta t)$. If not denoted explicitly in the following, we assume that $\Delta t$ is included in all records and lists.

\begin{algorithm}[t]
    \SetAlgoLined
    \DontPrintSemicolon
    Initialize feature array $X$ and labels $y$.\\
    \For{\text{patient record} $\mathcal{R}_j \in \mathcal{R}$}{
        \For{\text{visit time point} $t \in  \mathcal{T}^\text{visit}_j$}{
             // {For all follow up visits before the next medication adjustment:}\\
            
            $ \mathcal{T}^\text{follow-up}_j \leftarrow \{ t' \in         \mathcal{T}^\text{visit}_j | \ t' > t \text{ and }$ \\
          $\ \ \ \ \ \ \ \ \ \ \ \ \ \ \ \ \ \ \ \ \ \  \not\exists t_m \in \mathcal{T}^\text{med}_j \text{ s.t. } t \le t_m \le t'\}$\\
            \For{\text{follow up} $t' \in   \mathcal{T}^\text{follow-up}_j$}{
            $\Delta t \leftarrow t' - t$\\
            add $R_j(t, \Delta t)$ to $X$\\
            add $\text{score}_j(t + \Delta t) - \text{score}_j(t)$ to $y$
            }
        }
    }
    \caption{Sample Generation from Records $\mathcal{R}$}
    \label{alg:samplegen}
\end{algorithm}

\subsection{Sample Generation from Clinical Data}
To train ML models on a clinical dataset in a supervised fashion, input samples and the corresponding labels are constructed over all patient records $\mathcal{R}_j$ by iterating over the list of visit time points $\mathcal{T}^\text{visit}_j$ and over all follow-up visits until the next medication adjustment. The list of time points for medication treatments is denoted as $\mathcal{T}^\text{med}_j$. The sample generation procedure is shown in \Cref{alg:samplegen}.

\subsection{AdaptiveNet}

In order to deal with the above defined patient records $R_j(t)$ and the corresponding variable-sized event lists $E^{(\cdot)}(t)$ for a time point $t$, we propose the neural network architecture AdaptiveNet, which is able to deal with $K$ input sets $E^{1}, ..., E^{K}$, where every set can have variable length and different feature representations. With neural network modules $\phi^1, ..., \phi^K$, every element of the $K$ lists can be projected to a latent space and a sorted list of latent events can be computed as
$$ \Psi(R_j(t)) =  \Psi(E^{1}, ..., E^{K}) = sort \left( \bigcup_{k}  \bigcup_{e \in E^{ k}(t)}\phi^k(e)  \right),$$
with $1 \le k \le K$, sorted according to the time points of the events. The output vectors $\phi^k(\cdot) \in \mathbb{R}^F $ of the encoder networks $\phi^k$ have the same length $F$. These network modules can have an arbitrary architecture. In this work, we use fully-connected network modules to deal with numerical and categorical input features. We additionally propose to share the parameters of the last layer over all encoder networks. Then, $\phi^k(\cdot)$ can be seen as a projection of all input objects to the same encoded \textit{event space} (effects of which we investigate further in the results). The prediction $\hat y$ of the network is then computed as
$$\hat y(R_j(t))  = \rho\bigg( \text{LSTM} \bigg[  \Psi(R_j(t)) \bigg]  \ || \  e^\text{patient} \bigg),$$
where $||$ denotes concatenation of the vectors and $\rho$ is a fully-connected network module. In this work, we use an LSTM \cite{Hochreiter:1997:LSM:1246443.1246450} to pool the events by a recurrent unit. The scheme of the architecture is shown in \Cref{fig:adaptivenet}. To tackle the disease progression prediction problem, in this work, we use two encoder modules $\phi^1, \phi^2$ for the set of event types $\{\text{visit}, \text{med}\}$. It is straightforward to add other event types, for example imaging data. In this case, the encoder module could consist of a convolutional neural network, which can optionally be pre-trained and have fixed weights.

\section{Experimental Setup}
In this section, we first explain the dataset and describe how to perform disease progression prediction in RA. After that, we explain baselines and training details, such as hyperparameter optimization and architectural choices.

\subsection{Data Set}

In this work, we use the Swiss Clinical Quality Management (SCQM) database \cite{SCQM} for rheumatic diseases, which includes data of over 10.000 patients with RA, assessed during consultations and via the mySCQM online application. The database consists of general patient information, clinical data, disease characteristics, ultrasound, radiographs, lab values, medication treatments and patient reported outcome (HAQ, RADAI-5, SF12, EuroQol). Patients were followed-up with one to four visits yearly and clinical information was updated every time. The data collection was approved by a national review board and all individuals willing to participate, signing an informed consent form before enrolment in accordance with the Declaration of Helsinki.

\subsection{Disease Progression Prediction in RA}
To represent the disease level, we use the hybrid score DAS28-BSR as prediction target, which contains DAS28 (Disease Activity Score 28) and the inflammation bloodmarker BSR (blood sedimentation rate). DAS28 defines the disease activity based on 28 joints (number of swollen joints, number of painful joints and questionnaires).  For training and evaluation, we consider only visits with available DAS28-BSR score. We focus on 13 visit features, selected by a medical expert. All visit features are shown in \Cref{tab:visits}. Additionally, we consider eight medications. The corresponding features are listed in \Cref{tab:meds} and general patient features in \Cref{tab:patient}.

\begin{table}[t]
\small
    \centering
 \begin{tabular}{l c c}
    \toprule
   Numerical &  Missing     [\%]&  Mean  ($\pm$ Std.) \\
   \midrule
minimal disease activity & 1.6  &   1.3  $(\pm1.1)$  \\
number swollen joints& 6.6 &   3.3  $(\pm4.6)$  \\
number painful joints & 6.9  &   3.5 $(\pm5.3)$  \\
bsr & 14.6 &   18.5  $(\pm17.1)$  \\
das28bsr score & 16.4  &   3.2 $(\pm1.4)$  \\
pain level$^*$ & 22.4 &   3.3  $(\pm2.7)$  \\
disease activity &22.7  &   3.4 $(\pm2.7)$  \\
index$^*$ & & \\
haq score  & 27.8 &   0.8  $(\pm0.7)$  \\
weight [kg]  &36.5 &   70.7  $(\pm15.6)$  \\
height [cm]  & 40.8 &   165.3  $(\pm12.2)$  \\
crp  & 45.4 &   7.32  $(\pm12.7)$  \\
$\Delta t$ (5y history) & 0.0 &  2.2 $(\pm1.4)$\\ 
\midrule
Categorical&      & Values   [\%]\\
\midrule
morning  stiffness$^*$&  22.7 &  all day (1.9\%)\\
  & & $<$ 0.5h (15.4\%)\\
& & 0.5-1h  (12.0\%)\\
& &  $>$ 4h (1.6\%)\\
& & 12h (6.1\%)\\
& & 24h (3.5\%)\\
&  & no (36.8\%)\\
smoker  & 60.2 &  current (9.3\%)\\
& & former (12.3\%)\\
& & never (18.2\%)\\
\bottomrule
    \end{tabular}
    \caption{Visit features. The values for $\Delta t$ are shown for a prediction horizon of 1 year and a maximum history length of 5 years.  $(^*)$ This score is a rheumatoid arthritis disease activity index (RADAI).}
    \label{tab:visits}
\end{table}
\begin{table}[t]
\small
    \centering
 \begin{tabular}{l c c}
    \toprule
   Categorical & Value &  Pct. [\%] \\
   \midrule
drug & dmard mtx & 24.1 \\
    & prednison  & 16.8 \\
    & adalimumab & 7.9 \\
    & etanercept & 7.3 \\
    & tocilizumab & 4.0 \\
    & abatacept & 4.0 \\
    & rituximab & 3.5 \\
    & golimumab & 2.4 \\
    & other  & 30.1 \\
  type & prednison & 16.8 \\
         & dmard & 24.1 \\
         & biologic & 29.0\\
         & other  & 30.1\\
dose & no       & 41.3  \\
     &  $<$ 10 [mg]  &  9.6\\
    & 10 - 15 [mg]   &  12.6\\
    & $>$ 15 [mg]    &  36.5\\
\midrule
Numerical & Missing   [\%] & Mean  ($\pm$ Std.)\\
\midrule
$\Delta t$ (5y history) & 0.0 & 2.2  $(\pm 1.3)$ \\
    \bottomrule
    \end{tabular}
    \caption{Medication features. }
    \label{tab:meds}
    \vspace*{0.3cm }
     \begin{tabular}{l c c}
    \toprule
     Numerical & Missing   [\%] & Mean  ($\pm$ Std.)\\
\midrule
   age		     &  0.0	& 58.8 $(\pm 13.0)$ \\
 disease duration &  2.7     & 12.2 $(\pm 9.5 )$ \\
   \midrule
   Categorical &   &  Values [\%] \\
   \midrule
gender 	&  0.0    & male (26.0\%)\\
  		&    & female (74.0\%)\\
   r-factor & 9.1 & yes (62.9\%)\\
    		&       & no (28.0\%)\\
  anti-ccp & 31.6  & yes (42.4\%) \\
  			&   & no (26.0\%)\\
    \bottomrule
   
    \end{tabular}
    \caption{General patient features.}
    \label{tab:patient}
\end{table}

\subsection{Baselines}

As baselines, we consider a \textbf{Naive Baseline} where we set $f( R_j(t),\Delta t) \approx 0$, which means no change in the disease level. Further, we compare to a \textbf{Random Forest} (ensemble of decision trees) and a classical \textbf{Fully-Connected Network}. In order to deal with the variable patient history lengths of the clinical dataset, for both RF and the FCN architecture the inputs have to be padded with dummy values (e.g. $-1$). Since we want to consider the full patient history without any loss of information, we pad the input until the maximum number of visit features is reached for the patient with the longest history, and analogously for medication features. Considering a maximum history length over many years can lead to a huge input size and large amount of dummy values, complicating reliable estimation of relevant correlations considerably. For a history length of 5 years, the input sizes of the RF and FCN are 1178, due to a maximum of 35 visits for one patient. In contrast, AdaptiveNet has  8 input features for general patient information, 21 input features for visits and 18 features for medications. For a history length of 5 years, patients have in mean 6.3 $(\pm 5.3)$ visits and 2.5 $(\pm  2.7)$ medication adjustments.

\subsection{Training}

To train our models, we considered only patients with more than two visits and a minimum prediction horizon of 3 months up to a maximum of one year. In total, we trained on 28601 samples. For preprocessing, we scaled all features in the range $(0,1)$. The final architecture of AdaptiveNet is shown in \Cref{tab:adaptivenet}. As activation function, rectified linear units (ReLU) were used for all hidden layers. Additionally, we used $l1$-regularization for the weights of the network. For the FCN, we used dropout of 0.1, $l1$-regularization and three hidden layers of hidden dimension 32. For the RF, 100 tree estimators were used with a maximum depth of 12. To train the neural networks, weights $w$ are updated in supervised fashion via gradient descent on a minibatch of samples $X$ and labels $y$ of size $B$ using the update rule:
$$w \leftarrow w - \alpha \nabla_w L(w),$$
with learning rate $\alpha$ and loss $L$. The loss is computed as:
$$L(w) = \frac{1}{B} \sum_{i = 1}^{B} (y_i - \hat y(X_i))^2.$$

All architectures were trained with a batch size of $B=256$ for 7000 steps (batches of samples). For optimization, we used the Adam optimizer \cite{DBLP:journals/corr/KingmaB14} with a learning rate of $1e^{-4}$. To evaluate fairly, all models were optimized using grid search, including the baselines. The configuration spaces for all methods can be found in \Cref{tab:configspace}.

\begin{table}[t]

    \centering
\begin{tabular}{c}
       \toprule
      AdaptiveNet\\
        \midrule
              Input($B \times \text{seq}^\text{visit} \times 21$) and Input($B \times \text{seq}^\text{med} \times 18$) \\
        \midrule
         $\phi^\text{visit}$: FC($100$)$^{*}$,  $\phi^\text{med}$: FC($100$)$^{*}$  \\
        LSTM($\cdot$) \\
        \midrule
         concat($\cdot$, Input($B \times 8$))\\ 
       FC(100), FC(100)\\
       Linear(1)\\ 
         \bottomrule
    \end{tabular}
    \caption{Architecture of AdaptiveNet, where FC denotes fully-connected layers, $\text{seq}^{(\cdot)}$ variable-sized lists of events and $B$ the batch size. $(^{*})$ A second FC($100$) layer is used in experiments with parameter-sharing for the encoders.}
    \label{tab:adaptivenet}
\end{table}

\begin{table}[t]

    \centering
  \begin{tabular}{c|c |c}
        \toprule
        Model & Parameter     &  Config. Space\\
        \midrule
        RF & max depth   & [8, 10, \textbf{12}, 15]\\
            \midrule
        FCN & num hidden layers     & [2, \textbf{3}, 4]\\
            & hidden dim & [\textbf{32}, 64, 100]\\
            & dropout rate & [0.0, \textbf{0.1}, 0.25]\\
            \midrule
        AdaptiveNet & $\phi^{(\cdot)}$: hidden dim & [32, 64, \textbf{100}]\\
         & $\phi^{(\cdot)}$: num hidden layers & [\textbf{1},2]$^{*}$\\
         & $\rho$:  hidden dim & [64, \textbf{100}, 200]\\
         & $\rho$: num hidden layers & [1,\textbf{2}, 3]\\
            & dropout rate & [\textbf{0.0}, 0.1, 0.25]\\
        
        \bottomrule
        \end{tabular}
    \caption{Configuration spaces of the different approaches. The best performing architectural choices are shown in bold.  $(^{*})$ For experiments with parameter-sharing in the last layer, one additional layer of the same hidden dimension is added.}
    \label{tab:configspace}
\end{table}

\section{Results}
\label{sec:results}
\begin{figure}[t]
    \centering
    \includegraphics[width=0.45\textwidth]{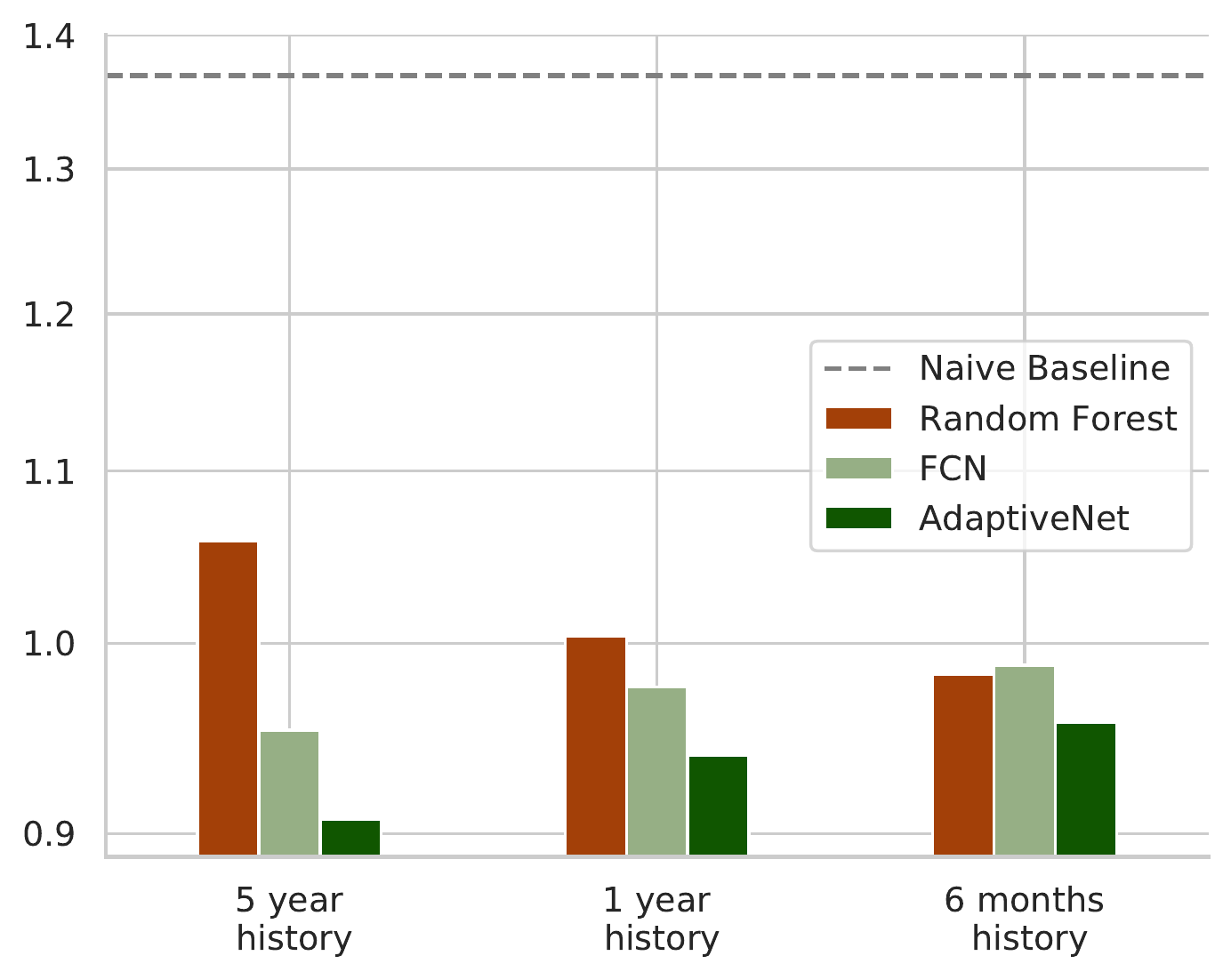}
    \caption{Mean squared error of the disease progression prediction for different maximum history lengths in a range from 5 years to 6 months. The prediction horizon is 1 year.}
    \label{fig:plot}
\end{figure}

To evaluate the performance of all models, we use 5-fold cross validation, where samples and corresponding labels were generated as shown in \Cref{alg:samplegen}. The results are shown in \Cref{fig:plot} for a prediction horizon of one year for all methods and different maximum history lengths of 6 months to 5 years.\\

All ML methods outperform the Naive Baseline significantly, which has a MSE of 1.369. The performance of the RF decreases with increasing history size from a MSE 0.983 for 6 months to a MSE of 1.058, probably due to the huge amount of input features. In contrast, both neural network architectures are able to profit from longer histories. With a MSE of 0.988 for history of 6 months, the FCN shows slightly worse performance than the RF. However, for longer histories, the FCN outperforms the RF, showing MSEs of 0.953 for 1 year and 1.058 for 5 years. The best performance over all history lengths is achieved by AdaptiveNet with a MSE of 0.957 for 6 months up to 0.907 for a history length of 5 years, which corresponds to an error of $7.94\%$ in the range of the target value (change of DAS28-BSR).\\

For further evaluation, we use the trained regression model to perform classification. Defining two classes as \textit{active disease} (DAS28-BSR $\le$ 2.6) and \textit{in remission} (DAS28-BSR $<$ 2.6), we can classify between future disease levels by estimating the absolute disease level with $f^\text{abs}(R_j(t), \Delta t) =  \text{score}_j(t) + f(R_j(t), \Delta t)$. AdaptiveNet achieves an accuracy of $76\%$ and an Area Under the Curve (AUC) score of $0.735$. Please note that these results can not directly be compared to the performance of the flare prediction approach shown in \cite{Vodencarevic} due to different definitions of active disease levels and different datasets (10.000 patients in this study vs. 314 patients).\\

\begin{figure}[t]
    \centering
    \includegraphics[width=0.45\textwidth]{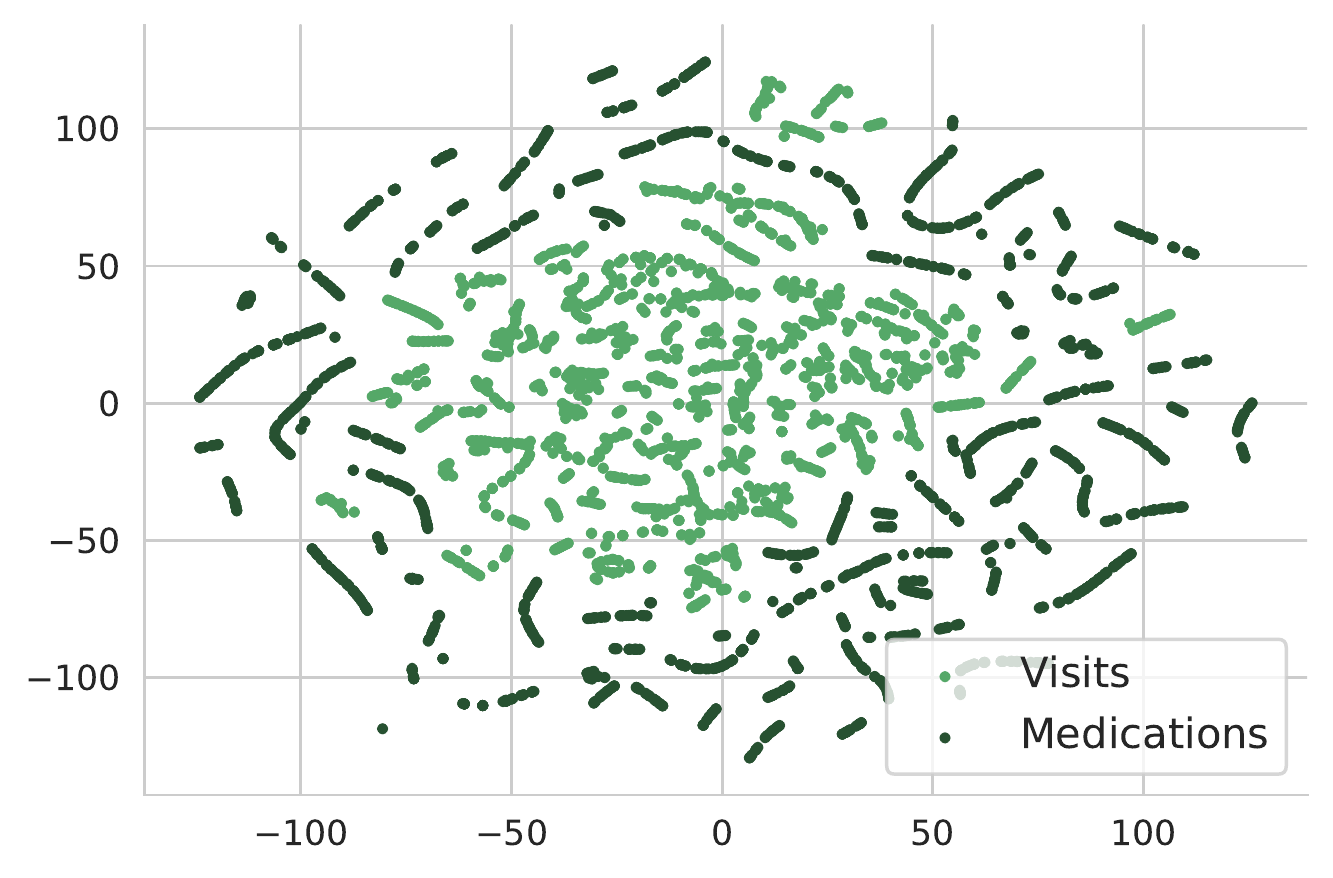}
    \caption{t-SNE visualization of the latent representations $\phi^\text{visit}(\cdot)$ and $\phi^\text{med}(\cdot)$ with shared parameters in the last layer.}
    \label{fig:tsne}
\end{figure}

Using parameter sharing in the last layer in the $\phi^{(\cdot)}$ modules, the network performs slightly better for long histories. With parameter sharing, the MSE of the disease progression prediction for AdaptiveNet for a considered maximum history of 5 years decreases from 0.923 to 0.899. The fact, that the individual latent representations for the different events are separated in our architecture makes it possible to characterize the structure of the encoded event space, which would not be possible for the other methods. We visualize the latent vectors $\phi^k(\cdot)$ using the t-distributed Stochastic Neighbor Embedding (t-SNE) \cite{vanDerMaaten2008} algorithm, which is a nonlinear dimensionality reduction method to reduce high-dimensional data to lower dimensions for visualization. \Cref{fig:tsne} shows the t-SNE plot for 6000 latent representations. As can be seen in the plot, the learned representations for visits and medications are nicely clustered and well separated, which partly explains the good performance of this architecture.

\section{Conclusion}

AdaptiveNet provides a flexible architecture to deal with multiple variable-sized lists of clinical events of different types, such as clinical visits or medication adjustments. The flexibility of the architecture allows to exploit and integrate \textit{all available data} into the decision making process in a unified and compact manner. Compared to classical approaches, AdaptiveNet is more robust in disease prediction, avoiding missing values and handling irregular visits by a recurrent unit. AdaptiveNet can be applied for various ML applications based on clinical datasets like EMRs, registries or trials. We are convinced that flexible and integrative deep learning systems such as AdaptiveNet can boost personalized medicine by considering as much data as possible to create high-quality machine-learned evidence.

\section*{Acknowledgement}
A list of rheumatology offices and hospitals that are contributing to the SCQM registries can be found on www.scqm.ch/institutions. The SCQM is financially supported by pharmaceutical industries and donors. A list of financial supporters can be found on www.scqm.ch/sponsor. We gratefully thank all patients and doctors for their support.

\balance
\bibliographystyle{aaai}
\bibliography{root}

\end{document}